\documentclass[letterpaper]{article} 
\usepackage{aaai25}
\usepackage{times}  
\usepackage{helvet}  
\usepackage{courier}  
\usepackage[hyphens]{url}  
\usepackage{graphicx} 
\usepackage{natbib}  
\usepackage{caption} 
\usepackage{algorithm}
\usepackage{algorithmic}
\usepackage{multirow}
\usepackage{array}
\usepackage{booktabs}
\usepackage{enumitem}
\usepackage{amsmath}
\usepackage{amssymb}
\usepackage[dvipsnames]{xcolor} 

\usepackage{colortbl}
\usepackage{pifont}
\definecolor{citecolor}{RGB}{34, 149, 34}
\usepackage[pagebackref=true,breaklinks=true,letterpaper=true,colorlinks,citecolor=black,bookmarks=false]{hyperref}

\usepackage{bm}

\usepackage{newfloat}
\usepackage{listings}
\DeclareCaptionStyle{ruled}{labelfont=normalfont,labelsep=colon,strut=off} 
\floatstyle{ruled}
\newfloat{listing}{tb}{lst}{}
\floatname{listing}{Listing}
\usepackage{xcolor}         
\newcommand{\MoU}{{\bf\texttt{MoU}}}

\newcommand{\MoAfull}{Mixture of Architectures}
\newcommand{\MoFfull}{Mixture of Feature Extractors}

\title{Mamba or Transformer for Time Series Forecasting? \\ Mixture of Universals (\MoU) Is All You Need}
\author{
    Sijia Peng$^1$,
    Yun Xiong$^1$\thanks{Corresponding author.},
    Yangyong Zhu$^1$,
    Zhiqiang Shen$^2$
}
\affiliations{
    \textsuperscript{\rm 1}Shanghai Key Laboratory of Data Science,  School of Computer Science, Fudan University\\

    \textsuperscript{\rm 2}Mohamed bin Zayed University of Artificial Intelligence\\
    sjpeng21@m.fudan.edu.cn, \{yunx, yyzhu\}@fudan.edu.cn, zhiqiang.shen@mbzuai.ac.ae
}

\begin{document}

\maketitle
\thispagestyle{plain}

\begin{abstract}

Time series forecasting requires balancing short-term and long-term dependencies for accurate predictions. Existing methods mainly focus on long-term dependency modeling, neglecting the complexities of short-term dynamics, which may hinder performance. 
Transformers are superior in modeling long-term dependencies but are criticized for their quadratic computational cost. Mamba provides a near-linear alternative but is reported less effective in time series long-term forecasting due to potential information loss. Current architectures fall short in offering both high efficiency and strong performance for long-term dependency modeling.
To address these challenges, we introduce Mixture of Universals (\MoU), a versatile model to capture both short-term and long-term dependencies for enhancing performance in time series forecasting. {\bf\texttt{MoU}} is composed of two novel designs: \MoFfull~(MoF), an adaptive method designed to improve time series patch representations for short-term dependency, and \MoAfull~(MoA), which hierarchically integrates Mamba, FeedForward, Convolution, and Self-Attention architectures in a specialized order to model long-term dependency from a hybrid perspective. The proposed approach achieves state-of-the-art performance while maintaining relatively low computational costs. Extensive experiments on seven real-world datasets demonstrate the superiority of {\bf\texttt{MoU}}. Code is available at \color{blue}{\url{https://github.com/lunaaa95/mou/}.}

\end{abstract}

%

\section{Introduction}
Time series forecasting is crucial in various fields, such as climate prediction~\cite{murat2018forecasting,scher2020artificial,haq2022cdlstm,neumann2024intrinsic}, financial investment~\cite{sezer2020financial,liu2023financial,bieganowski2024supervised}, and household power management~\cite{bilal2022comparative,kim2023time,cascone2023predicting}. However, achieving accurate forecasts is challenging as both short-term and long-term dynamics jointly influence future values~\cite{chen2023multi}.

\begin{figure}[t]
    \centering
    \includegraphics[width=0.99\linewidth]{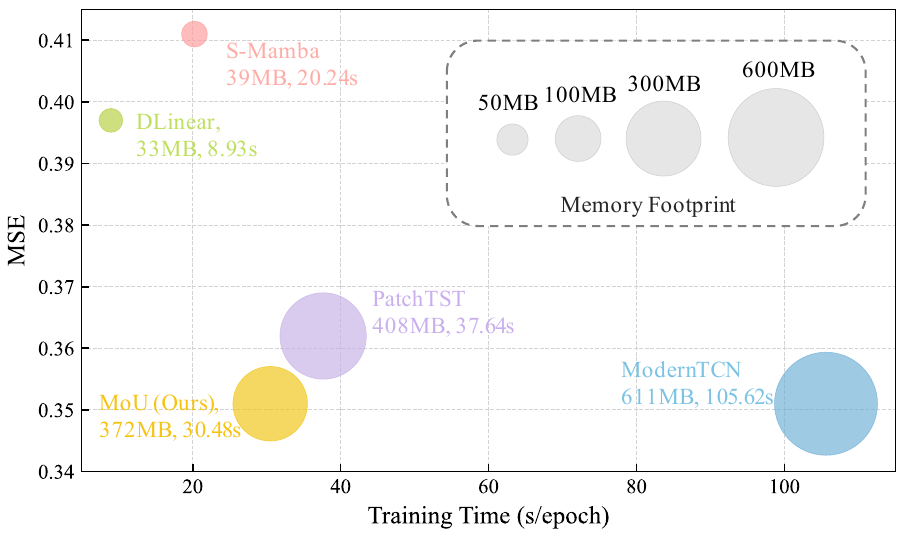}
    \vspace{-0.12in}
    \caption{Model efficiency comparison. The results are on ETTm2 with forecasting length of 720 by a unified testing.}
    \label{fig:efficiency}
    \vspace{-0.15in}
\end{figure}

To model short-term dependency, the patching strategy has gained attention for its ability to preserve contextual semantics. For instance, PatchTST~\cite{nie2022time} proposes grouping data points into patch tokens to retain local semantics, while Pathformer~\cite{chenpathformer} employs a multi-scaled patching strategy to summarize short-term dependency in various resolutions. Although these approaches enhance short-term information by generating patch tokens, they still rely on uniform linear transformations for patch embedding. This approach neglects the divergence in feature affinity within different patches, which arises from their varying semantic contexts. As a result, important contextual information may be lost, limiting the accurate representation of short-term details.

\begin{figure*}[t]
    \centering
    \includegraphics[width=0.95\linewidth]{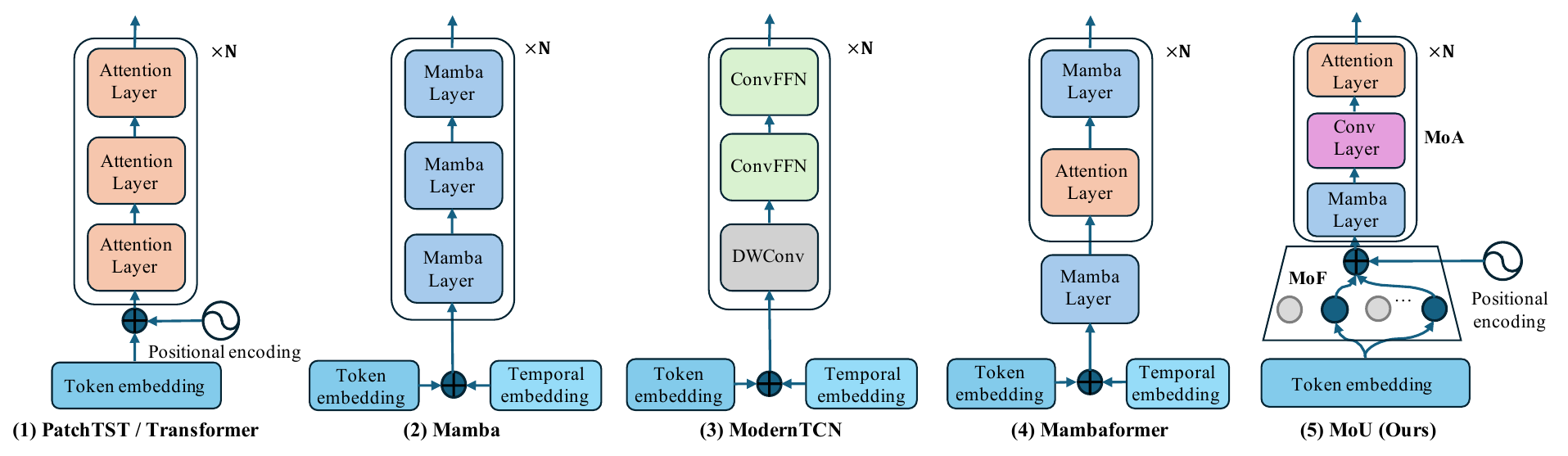}
    \vspace{-0.12in}
    \caption{Illustration of different architectures for long-term time series forecasting. From left to right are PatchTST / Transformer~\cite{nie2022time}, Mamba~\cite{gu2023mamba,wang2024mamba}, ModernTCN~\cite{donghao2024moderntcn}, Mambaformer~\cite{xu2024integrating}, and our proposed \MoU. Feed-forward layer is omitted for simplicity in Transformer and our model.}
    \label{fig:diff}
    \vspace{-0.15in}
\end{figure*}

This motivates us to explore adaptive methods that tailor the embedding process for different patch tokens, generating more informative representations. Such approaches have shown success in computer vision \cite{chen2020dynamic, yang2019condconv}. For instance, Dynamic Convolution~\cite{chen2020dynamic} and Conditional Convolution~\cite{yang2019condconv} adapt convolution kernel parameters for different input patches. However, we find that these methods perform poorly for time series patches, even worse than uniform linear transformation (for detailed explanation, please refer to Section \ref{sec:ablation}). We attribute this failure to the increased tunable parameters relative to the small size of patch data, which may hinder the feature extractor's ability to learn robust representations. 

Given the lack of suitable adaptive methods for time series patches, we propose \MoFfull~(MoF). Inspired by the Mixture of Experts approach~\cite{shazeer2017outrageously}, which uses sparse activation to flexibly adjust model structure for various downstream tasks, MoF comprises multiple sub-extractors designed to handle divergent contexts within patches. With sparse activation, MoF selectively activates the most appropriate sub-extractor based on the input patch, ensuring both the learning of diverse contexts and minimal parameter increase. 
By effectively capturing the varying contextual information of different patches, MoF stands as a promising feature extractor for time series data.

For capturing long-term dynamics, Transformers are effective due to their global awareness via the attention mechanism \cite{vaswani2017attention}. However, they suffer from quadratic computational costs in each Self-Attention layer, which are often stacked multiple times \cite{nie2022time, chen2024multi}. Mamba is recently proposed for long-term sequence modeling with near-linear computational cost \cite{gu2023mamba}, but it has shown limited performance in long-term time series forecasting, likely due to information loss from its compression and selection mechanism \cite{wang2024mamba}. To address both effectiveness and efficiency, we introduce \MoAfull~ (MoA), a novel hybrid encoder for time series long-term forecasting. MoA features a hierarchical structure starting with a Mamba layer that selects and learns key dependencies using a Selective State-Space Model (SSM). This is followed by a FeedForward transition layer and a Convolution-layer that broadens the receptive field to capture longer dependencies. Finally, a Self-Attention layer integrates information globally to fully capture long-term dependencies. By initially focusing on partial dependencies and progressively expanding to a global view, MoA achieves better long-term dependency learning with minimal cost. Additionally, using the Self-Attention layer only once reduces the computational burden. Efficiency and model size comparisons are shown in Figure~\ref{fig:efficiency}.

Our key contributions of this work are as follows:
\begin{itemize}
 
 \item We propose Mixture of Feature Extractors (MoF) as an adaptive feature extractor that captures the contextual semantics of patches to enhance short-term representation. To our knowledge, MoF is the first adaptive method for embedding time series patches. We also introduce Mixture of Architectures (MoA) for hierarchical long-term dependency modeling.
 \item We introduce {\bf\texttt{MoU}}, integrating the adaptive feature extraction capabilities of MoF with a comprehensive long-term dependency modeling of MoA. It pioneers the decomposition of long-term dependencies from a partial-to-global view and applies such a diverse mixture to time series forecasting. Architecture comparison of our {\bf\texttt{MoU}} is provided in Figure~\ref{fig:diff}.
 \item We conduct extensive experiments on seven real-world datasets to evaluate the performance of {\bf\texttt{MoU}} in time series long-term forecasting task, the results show that {\bf\texttt{MoU}} consistently achieves state-of-the-art results on the majority of the datasets. 
\end{itemize}

\section{Approach}

\subsection{Problem Setting and Model Structure}

The aim of multivariate time series forecasting is to predict future values over $T$ time steps, given historical data from the past $L$ time steps. Specifically, given the historical values of a multivariate time series consisting of $M$ variables, ${\bf{X}}_\text{input} = \left[{\bf X}^1, {\bf X}^2, \ldots, {\bf X}^M\right] \in \mathbb{R}^{M \times L}$, where each ${\bf X}^i$ is a vector of length $L$, ${\bf X}^i = \left[\bm x_1, \bm x_2, \ldots, \bm x_L\right] \in \mathbb{R}^{L}$, our task is to predict the future values $\hat{\bf X}_\text{output} = \left[\hat{\bf X}^1, \hat{\bf X}^2, \ldots, \hat{\bf X}^M\right] \in \mathbb{R}^{M \times T}$, where each $\hat{\bf X}^i$ is a vector of length $T$, representing the predicted values for the $i$-th variable from $L+1$ to $L+T$. In this work, we adopt the variate independence setting as in PatchTST~\cite{nie2022time}, simplifying our goal to learning a function $\mathcal{F}$: $\bf X \rightarrow \hat{\bf X}$, which maps historical time series to predicted time series. In our work, $\mathcal{F}$ is our proposed {\bf\texttt{MoU}}.

First, we preprocess the time series using a patching strategy, which applies a sliding window of fixed size $P$ with stride $S$ to generate a sequence of $N$ patch tokens:
\begin{equation}
\label{eq:patch}
{\bf X}_p = \operatorname{Patch}(\bf X)
\end{equation}
where ${\bf X} \in \mathbb{R}^{L}$ is the raw time series, and ${\bf X}_p \in \mathbb{R}^{N \times P}$ is the resulting patched sequence, consisting of $N$ patch tokens, each containing $P$ data points.

To capture short-term dependencies within tokens, $X_p$ is fed into our MoF module. The process of generating adaptive representations is described as:
\begin{equation}
\label{eq:mof}
    {\bf X}_\text{rep} = \operatorname{MoF}({\bf X}_p),
\end{equation}
where $\operatorname{MoF}(\cdot)$ is our proposed Mixture of Feature Extractors, specifically designed for time series patch tokens. It adjusts network parameters for different patches based on their divergent context information while maintaining low computational cost. The detailed description of $\operatorname{MoF}(\cdot)$ will be presented in Section \ref{sec:mof}. After applying $\operatorname{MoF}(\cdot)$, we obtain $\bf X_\text{rep} \in \mathbb{R}^{N \times D}$, where $D$ denotes the feature dimension of the patch tokens.

To capture long-term dependencies among tokens, $X_\text{rep}$ is processed by our proposed MoA. MoA uses multiple layers of diverse architectures to model long-term dependencies from with a hybrid perspective:
\begin{equation}
\label{eq:moa}
   {\bf X}_\text{rep'} = \operatorname{MoA}(\bf X_\text{rep}),
\end{equation}
where $ \operatorname{MoA}(\cdot)$ denotes our long-term encoder based on the Mixture of Architectures. Its detailed description will be provided in Section \ref{sec:moa}. This step yields $X_\text{rep'} \in \mathbb{R}^{N \times D}$.

Finally, we flatten $X_\text{rep'}$ and apply a linear projector to obtain the final prediction:
\begin{equation}
\label{eq:head}
    \hat{\bf X} = \mathbf{P}(\operatorname{Flatten}(\bf X_\text{rep'})).
\end{equation}

\begin{figure}[t]
    \centering
    \includegraphics[width=0.99\linewidth]{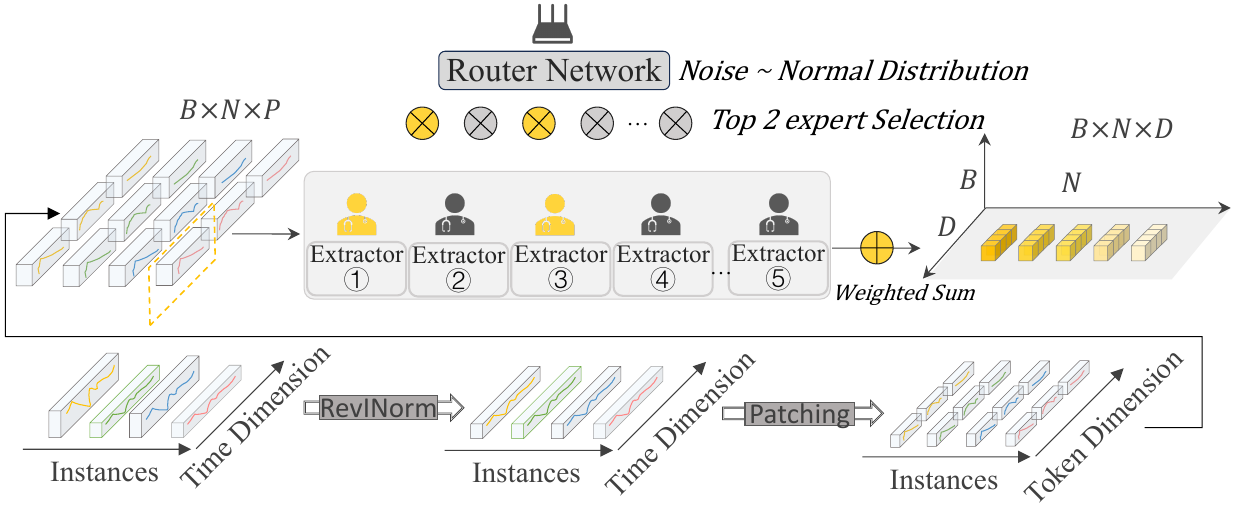}
    \vspace{-0.12in}
    \caption{Illustration of the proposed Mixture of Feature Extractors (MoF) structure. MoF contains multiple Sub-Extractors, each is tailored to learn different contexts within individual patches. Sub-Extractors are selectively activated by Router in a sparse manner, thereby ensuring both adaptivity and high efficiency.}
    \label{fig:mof}
    \vspace{-0.15in}
\end{figure}

\subsection{\MoFfull}
\label{sec:mof}
To account for the diverse feature affinities present within various patches, we introduce MoF in Figure~\ref{fig:mof} as an adaptive feature extractor specifically designed for time series patches. Different from other adaptive methods, MoF keeps minimum increment in activated net parameters by sparse activation. This facilitates a robust generation of adaptive representations, especially for time series patches with such small data size.
Specifically, MoF consists of a set of sub-extractors $\{F_1, F_2, \ldots, F_c\}$, each representing an independent linear mapping. The representation for a patch token is generated by MoF as follows:
\vspace{-0.1in}
\begin{equation}
 {\bf X}_\text{rep}=\operatorname{MoF}({\bf X}_p)=\sum_{i=1}^{n}R_i({\bf X}_p)F_i({\bf X}_p)
\end{equation}
where $R_i(\cdot)$ is an input-relevant router that generates a sparse vector with most elements set to zero, enabling the sparse activation of sub-extractors and ensuring minimal parameter increase. The router function $R({\bf X}_p)_i$ is calculated:
\begin{equation}
R({\bf X}_p)_i=\operatorname{Softmax}(\operatorname{Top_k}(H({\bf X}_p)_i, k))
\end{equation}
where $\operatorname{Softmax}(\cdot)$ normalizes the top $k$ scores kept by $\operatorname{Top_k}(\cdot, k)$. $H({\bf X}_p)$ is a vector of scores for sub-extractors: $H({\bf X}_p)=[H({\bf X}_p)_1, H({\bf X}_p)_2,...H({\bf X}_p)_c]$, where $H({\bf X}_p)_i$ denotes the score of $i$-th sub-extractor:
\begin{equation}
H({\bf X}_p)_i \!=\! \left({\bf X}_p \!\cdot\! W_g\right)_i   \!+\! \operatorname{SN} \!\cdot\! \operatorname{Softplus}\left(\left({\bf X}_p \!\cdot\! W_{\text{noise}}\right)_i\right)
\end{equation}
where $W_g$ contains the parameters of a linear function, and the second term injects tunable noise for load balancing, following the approach in MoE~\cite{shazeer2017outrageously}. Here, SN represents the standard normal distribution.

This mechanism effectively partitions the patch tokens into combinations of $c$ different patterns, where $c$ corresponds to the number of sub-extractors. Since the information of each pattern is processed by corresponding optimal feature extractor, MoF is capable of generating most representative embedding for patches with divergent contexts.

\begin{figure}[t]
    \centering
    \includegraphics[width=0.99\linewidth]{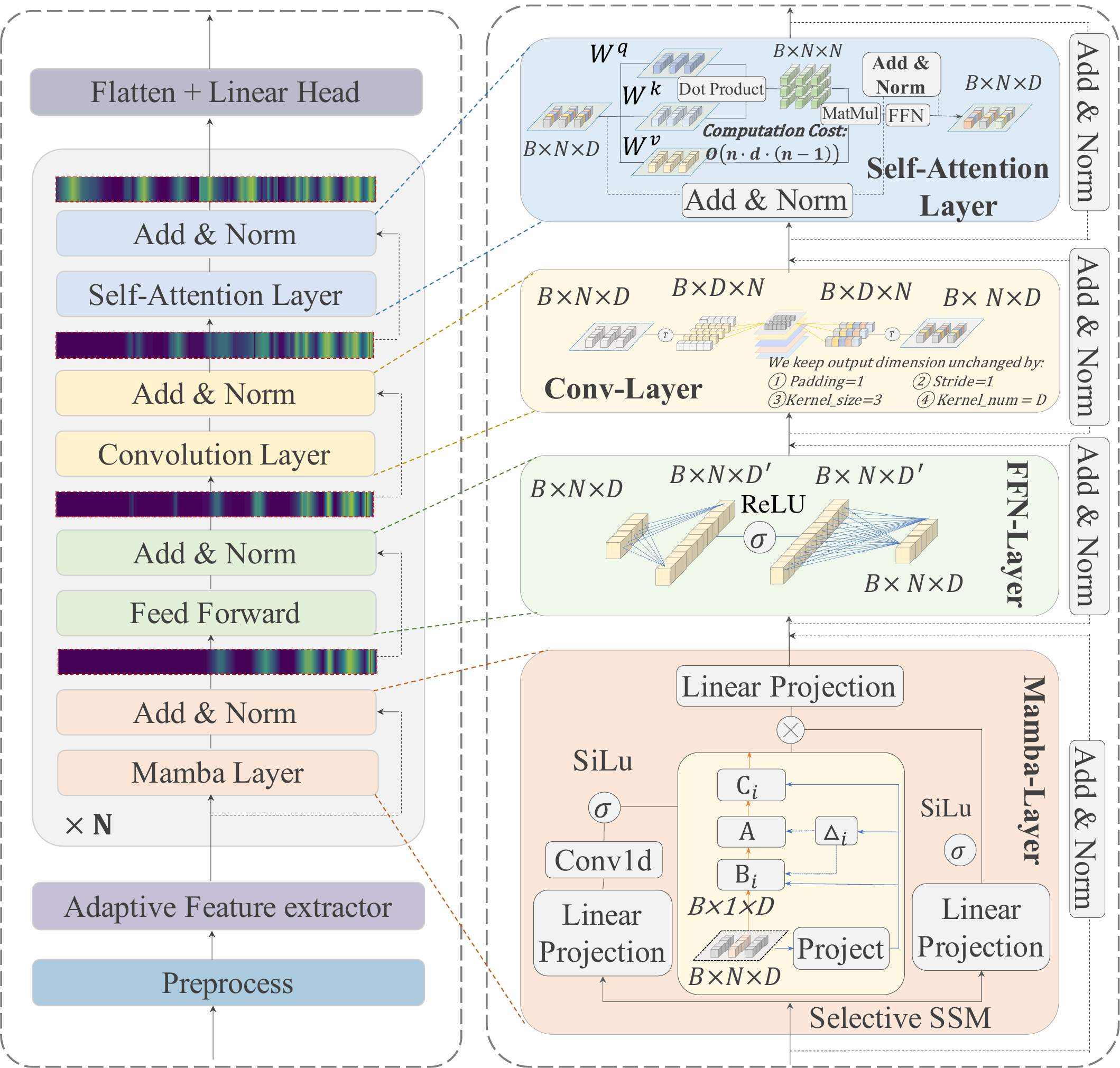}
    \vspace{-0.12in}
    \caption{Illustration of the proposed Mixture of Architectures (MoA) structure. MoA initially concentrates on part of dependencies selected by SSM, and progressively expands the receptive field into a comprehensive global view. }
    \label{fig:moa}
    \vspace{-0.15in}
\end{figure}

\subsection{\MoAfull}
\label{sec:moa}
We propose MoA to capture comprehensive long-term dependencies. As shown in Figure~\ref{fig:moa}, MoA is structured hierarchically with four layers, Mamba, FeedForward, Convolution, and Self-Attention layers, each captures a different aspect of long-term dependencies. Benefiting from its gradually expanding perspective, MoA is capable of modeling long-term dependencies with effectiveness and efficiency.

\noindent\textbf{Mamba-layer in Time Series} is first to select relevant data and learns time-variant dependencies with selective SSM. 
Let ${\bm x}$ denotes the ${\bf X}_\text{rep}$, the process can be described as:
\begin{equation}
\begin{gathered}
{\bm x}'=\sigma\left(\operatorname{Conv1D}\left(\operatorname{Linear}\left({\bm x}\right)\right)\right) \\
{\bm z}=\sigma\left(\operatorname{Linear}\left({\bm x}\right)\right)
\end{gathered}
\end{equation}
where $\sigma$ is activation function $\operatorname{SiLU}$. Then, ${\bm x}$, ${\bm x}'$ and ${\bm z}$ are used to calculate output $y$ by:

\begin{equation}
\begin{gathered}
\left.{\bm y}'=\operatorname{Linear}(\operatorname{SelectiveSSM}({\bm x}') \otimes {\bm z}\right)
\\
{\bm y}=\operatorname{LayerNorm}\left({\bm y}'+{\bm x}\right)
\end{gathered}
\end{equation}
where $\otimes$ denotes element-wise multiplication, and the $\operatorname{SelectiveSSM}$ can be further expanded as:
\begin{equation}
\begin{gathered}
\operatorname{SelectiveSSM}({\bm x}'_t) = {\bm y}_t \\
{\bm y}_t=C {\bm h}_t, \quad {\bm h_t}=\bar{A} {\bm h}_{t-1}+\bar{B} {\bm x}'_t
\end{gathered}
\end{equation}
Specifically, $h_t$ is the latent state updated at time step t, while $y_t$ is the output representation. The discrete matrices $\overline{A}$, $\overline{B}$ and $C$ are input-relevant and updated by the time-variant recurrent rule over the time:
\begin{equation}
\begin{gathered}
B_t=S_B\left({\bm x}'_t\right), \quad C_t=S_C\left({\bm x}'_t\right), \\ \Delta_t=\operatorname{softplus}\left(S_{\Delta}\left({\bm x}'_t\right)\right) \\
\end{gathered}
\end{equation}
where $S_B$, $S_C$, and $S_\Delta$ are linear projection layers. We use following equation to obtain discrete parameters $\bar{A}_t$ and $\bar{B}_t$:
\begin{equation}
\label{eq:m_para}
\begin{gathered}
    f_A\left(\Delta_t, A\right)=\exp \left(\Delta_t A\right)\\
    f_B\left(\Delta_t, A, B_t\right)=(\Delta_t A)^{-1}(\exp (\Delta_t A)-I) \cdot \Delta B_t \\
\bar{A}_t=f_A\left(\Delta_t, A\right), \quad \bar{B}_t=f_B\left(\Delta_t, A, B_t\right)
\end{gathered}
\end{equation}
where $f_A$ and $f_B$ are discretization functions, while $A$,$B$,$C$ and $\Delta$ are parameters.

\noindent\textbf{FeedForward-layer} serves as the second transition layer, following the Transformer architecture's convention of enhancing non-linearity:
   $ {\bm x}_\text{ffn}=\operatorname{FeedForward}({\bm y}_t\;; w_1,\sigma, w_2) $, 
where $w_1$ and $w_2$ are parameters, $\sigma$ is activation function.

\noindent\textbf{Convolution-layer}, the third layer, expands MoA’s receptive field. It facilitates information exchange among tokens encapsulating partial long-term dependencies, allowing more comprehensive learning of dependencies: 
   ${\bm x}_\text{conv}=\operatorname{Conv}({\bm x}_\text{ffn}; {\bf k}, s, p, c_\text{out}) $, 
where ${\bf k}$ is the kernel size, $s$ is the stride, $p$ is the padding, and $c_\text{out}$ is the number of output channels, with the output dimension kept unchanged.

\noindent\textbf{Self-Attention-layer} is the final layer, capturing comprehensive long-term dependencies with its global perspective: 

\begin{equation}
\begin{gathered}
\bm{x}_{\text{att}}=\operatorname{FeedForward}(\operatorname{Attention}(Q, K, V))\\
\operatorname{Attention}(Q, K, V) = \operatorname{Softmax} \left( \frac{QK^T}{\sqrt{d_k}} \right)V \\
Q = \bm{x}_\text{conv} W_Q, \quad K = \bm{x}_\text{conv} W_K, \quad V = \bm{x}_\text{conv} W_V
\end{gathered}
\end{equation}
where $W_Q$,$W_K$,$W_V$ are parameters, and self-attention is implemented using a scaled dot product.

\noindent{\bf Partial-to-global Design for Time Series.} 
To capture long-term dependencies in time series, our MoA initially focusing on part of dependencies and gradually expanding into a comprehensive global perspective. It begins with the Mamba-layer, which selectively processes time-variant dependencies using SSM, focusing on partial dependencies to identify essential features. The FeedForward-layer adds non-linearity and transitions these partial dependencies into more complex representations. The Convolution-layer then broadens the receptive field, facilitating information exchange between tokens to enhance understanding of broader temporal relationships. Finally, the Self-Attention-layer provides a global perspective, integrating localized information into a complete understanding of long-term dependencies. This hierarchical design ensures the the capturing of intricate time series patterns while remaining computationally efficient.

\begin{table*}[!t]
\centering
\caption{Multivariate long-term forecasting results with \MoU~on seven real-world datasets. We use prediction length \( T \in \{24, 36, 48, 60\} \) for ILI dataset and \( T \in \{96, 192, 336, 720\} \) for others. Best results are shown in \textbf{bold}, second best results are shown in \underline{underline}.}
\vspace{-0.1in}
\label{t:main_result}
\resizebox{0.75\textwidth}{!}{
\begin{tabular}{cc|cc|cc|cc|cc|cc}
\toprule
\multicolumn{2}{c|}{Model} & \multicolumn{2}{c|}{\MoU~ (Ours)} & \multicolumn{2}{c|}{ModernTCN} & \multicolumn{2}{c|}{PatchTST} & \multicolumn{2}{c}{DLinear} & \multicolumn{2}{c}{S-Mamba}\\ 
\hline
\multicolumn{2}{c|}{Metric} & \multicolumn{1}{c}{MSE} & \multicolumn{1}{c|}{MAE} & \multicolumn{1}{c}{MSE} & \multicolumn{1}{c|}{MAE} & MSE & MAE & MSE & MAE & MSE & MAE\\ 

\hline
\multirow{4}{*}{\rotatebox{90}{ETTh1}} 
& 96 & \textbf{0.358} & \textbf{0.393} & \underline{0.368} & \underline{0.394} & 0.370 & 0.400 & 0.375 & 0.399 & 0.398 & 0.417 \\
& 192 & \textbf{0.402} & 0.418 & 0.406 & \textbf{0.414} & 0.413 & 0.429 & \underline{0.405} & \underline{0.416} & 0.438 & 0.444 \\
& 336 & \textbf{0.389} & \underline{0.418} & \underline{0.392} & \textbf{0.412} & 0.422 & 0.440 & 0.439 & 0.443 & 0.453 & 0.455 \\
& 720 & \textbf{0.440} & \underline{0.462} & 0.450 & \textbf{0.461} & \underline{0.448} & 0.468 & 0.472 & 0.490 & 0.510 & 0.508 \\
 \hline
 
\multirow{4}{*}{\rotatebox{90}{ETTh2}} 
& 96 & \underline{0.266} & \textbf{0.332} & \textbf{0.264} & \underline{0.333} & 0.274 & 0.336 & 0.289 & 0.353 & 0.301 & 0.358 \\
& 192 & \underline{0.319} & \textbf{0.369} & \textbf{0.318} & \underline{0.373} & 0.339 & 0.379 & 0.383 & 0.418 & 0.366 & 0.398 \\
& 336 & \textbf{0.305} & \textbf{0.369} & \underline{0.314} & \underline{0.376} & 0.331 & 0.380 & 0.448 & 0.465 & 0.391 & 0.419 \\
& 720 & \textbf{0.379} & \textbf{0.422} & 0.394 & 0.432 & \textbf{0.379} & \textbf{0.422} & 0.605 & 0.551 & 0.417 & 0.444 \\
 \hline
 
\multirow{4}{*}{\rotatebox{90}{ETTm1}} 
& 96 & \textbf{0.292} & 0.346 & 0.297 & 0.348 & \underline{0.293} & \textbf{0.343} & 0.299 & \textbf{0.343} & 0.303 & 0.358 \\
& 192 & \textbf{0.329} & 0.371 & 0.334 & \underline{0.370} & \underline{0.330} & \textbf{0.368} & 0.335 & 0.365 & 0.345 & 0.383 \\
& 336 & \textbf{0.358} & \underline{0.391} & 0.370 & 0.392 & \underline{0.365} & 0.392 & 0.369 & \textbf{0.386} & 0.378 & 0.403 \\
& 720 & \textbf{0.412} & 0.420 & \underline{0.413} & \textbf{0.416} & \underline{0.419} & 0.424 & 0.425 & 0.421 & 0.440 & 0.440 \\
 \hline
 
\multirow{4}{*}{\rotatebox{90}{ETTm2}} 
& 96 & \textbf{0.166} & 0.257 & 0.169 & \textbf{0.256} & \textbf{0.166} & \textbf{0.256} & 0.167 & 0.260 & 0.177 & 0.270 \\
& 192 & \textbf{0.220} & \textbf{0.294} & 0.227 & 0.299 & \underline{0.223} & \underline{0.296} & 0.224 & 0.303 & 0.229 & 0.305 \\
& 336 & \textbf{0.270} & \textbf{0.329} & 0.276 & \textbf{0.329} & \underline{0.274} & 0.330 & 0.281 & 0.342 & 0.281 & 0.338 \\
& 720 & \textbf{0.351} & \textbf{0.380} & \textbf{0.351} & \underline{0.381} & 0.362 & 0.385 & 0.397 & 0.421 & 0.371 & 0.392 \\
 \hline
 
\multirow{4}{*}{\rotatebox{90}{Weather}} 
& 96 & \textbf{0.149} & \underline{0.199} & 0.150 & 0.204 & \textbf{0.149} & \textbf{0.198} & 0.152 & 0.237 & 0.158 & 0.210 \\
& 192 & \textbf{0.193} & \textbf{0.241} & 0.196 & 0.247 & \underline{0.194} & \textbf{0.242} & 0.220 & 0.282 & 0.202 & 0.251 \\
& 336 & \textbf{0.234} & \textbf{0.279} & \underline{0.237} & 0.283 & 0.245 & \underline{0.282} & 0.265 & 0.319 & 0.256 & 0.291 \\
& 720 & \textbf{0.308} & \textbf{0.329} & 0.315 & 0.335 & \underline{0.314} & \underline{0.333} & 0.323 & 0.362 & 0.329 & 0.341 \\
 \hline
 
\multirow{4}{*}{\rotatebox{90}{Illness}} 
& 24 & 1.468 & 0.774 & \underline{1.367} & \textbf{0.720} & \textbf{1.301} & \underline{0.734} & 2.215 & 1.081 & {1.995} & {0.873} \\
& 36 & \textbf{1.269} & \textbf{0.692} & \underline{1.345} & \underline{0.760} & 1.658 & 0.898 & 1.963 & 0.963 & {1.963} & {0.864} \\
& 48 & \underline{1.613} & \textbf{0.827} & \textbf{1.526} & \underline{0.837} & 1.657 & 0.879 & 2.130 & 1.024 & {1.773} & {0.863} \\
& 60 & \underline{1.650} & \underline{0.841} & 1.838 & 0.878 & \textbf{1.436} & \textbf{0.790} & 2.368 & 1.096 & {2.176} & {0.960} \\
 \hline
 
\multirow{4}{*}{\rotatebox{90}{Electricity}} 
& 96 & \textbf{0.127} & \textbf{0.222} & 0.131 & 0.226 & \underline{0.129} & \textbf{0.222} & 0.153 & 0.237 & 0.135 & 0.231 \\
& 192 & \underline{0.145} & \textbf{0.238} & \textbf{0.144} & \underline{0.239} & 0.147 & 0.240 & 0.152 & 0.249 & 0.157 & 0.252 \\
& 336 & \underline{0.163} & 0.262 & \textbf{0.161} & \textbf{0.259} & \underline{0.163} & \textbf{0.259} & 0.169 & 0.267 & 0.175 & 0.271 \\
& 720 & \underline{0.193} & \underline{0.290} & \textbf{0.191} & \textbf{0.286} & 0.197 & \underline{0.290} & 0.233 & 0.344 & 0.196 & 0.293 \\
 \midrule
 \multicolumn{2}{c|}{Win count} & \multicolumn{2}{c|}{\color{red}{\textbf {35}}} & \multicolumn{2}{c|}{\underline{17}} & \multicolumn{2}{c|}{14} & \multicolumn{2}{c|}{2} & \multicolumn{2}{c}{0}\\ 
\bottomrule
 
\end{tabular}
}
\end{table*}

\subsection{Computational Complexity and Model Parameter}
\label{sec:complexity}

Given a sequence with $T$ tokens, the computational cost of an \MoU~block that selects the top-$k$ experts is:
\begin{equation}
    C_{\MoU} \!=\underbrace{kT\!\times\! d^2}_\text{MoF} \!+\! \underbrace{T \!\times\! d^2}_\text{Mamba}+ \underbrace{T\!\times\! d^2}_\text{FFN} + \underbrace{{\bf k}Td^2}_\text{Conv} \!+\! \underbrace{T^2 \!\times\! d + T \!\times\! d^2}_\text{Transformer}
\end{equation}
where $\bf k$ is the kernel size in the convolutional layer, $d$ is the dimension of the vector representations. In the Transformer block, we account for the complexity of both the linear transformation used to compute  query ($Q$), key ($K$), and value ($V$) matrices, as well as the complexity of the Self-Attention layer. For comparison, the computational cost required by a three-layer Multi-Head Self-Attention (MHSA) is:
\begin{equation}
    C_\text{MHSA} = 3\times(T^2d+Td^2)
\end{equation}
As shown, except for the Transformer block, the complexity of our structure is linear, resulting in significantly lower computational requirements compared to pure Transformer models like PatchTST~\cite{nie2022time}. Considering the model parameters, the linear layer primarily involves parameters of size $d^2 + \text{bias}$ for MoF and FFN layers, the convolutional layer is mainly determined by the kernel size of ${\bf k} \times d^2$, and the Self-Attention layer's parameters are mainly the mapping matrices for $Q, K, V$, each head of size $d^2$. As shown in Figure~\ref{fig:efficiency}, the total size of our model is 372MB, compared to PatchTST of 408MB and ModernTCN of 611MB.

\section{Experiments}

\begin{table*}[tp]
\centering
\caption{Ablation experiments for feature extractors in modeling short-term dependencies. We compare three types of adaptive feature extractors and one uniform feature extractor. Best results are shown in \textbf{bold}.}
\label{t:abation1}
\vspace{-0.1in}
\resizebox{0.6\textwidth}{!}{
\begin{tabular}{cc|cc|cc|cc|cc}
\toprule
\multicolumn{2}{c|}{Model} & \multicolumn{2}{c|}{MoF (in \MoU~)} & \multicolumn{2}{c|}{SE-M} & \multicolumn{2}{c|}{Linear} & \multicolumn{2}{c}{Dyconv}\\ 
\hline
\multicolumn{2}{c|}{Metric} & \multicolumn{1}{c}{MSE} & \multicolumn{1}{c|}{MAE} & \multicolumn{1}{c}{MSE} & \multicolumn{1}{c|}{MAE} & MSE & MAE & MSE & MAE\\ 

\hline
\multirow{4}{*}{\rotatebox{90}{ETTh1}} 
& 96 & \textbf{0.358} & \textbf{0.393} & 0.363 & 0.397 & 0.361 & 0.395 & 0.468 & 0.460 \\
& 192 & \textbf{0.402} & \textbf{0.418} & 0.408 & 0.423 & 0.405 & 0.421 & 0.497 & 0.475 \\
& 336 & \textbf{0.389} & \textbf{0.418} & 0.394 & 0.424 & 0.394 & 0.424 & 0.467 & 0.474 \\
& 720 & \textbf{0.440} & \textbf{0.462} & 0.486 & 0.465 & 0.441 & 0.466 & 0.504 & 0.506 \\
 \hline
 
\multirow{4}{*}{\rotatebox{90}{ETTm2}} 
& 96 & 0.166 & 0.257 & \textbf{0.165} & \textbf{0.256} & 0.166 & 0.257 & 0.206 & 0.294 \\
& 192 & \textbf{0.220} & \textbf{0.294} & 0.223 & 0.297 & \textbf{0.220} & 0.295 & 0.254 & 0.324 \\
& 336 & 0.270 & \textbf{0.329} & \textbf{0.268} & 0.330 & 0.273 & 0.332 & 0.301 & 0.350 \\
& 720 & \textbf{0.351} & \textbf{0.380} & 0.356 & 0.381 & 0.353 & 0.381 & 0.383 & 0.400 \\
 \hline
 
\multirow{4}{*}{\rotatebox{90}{Weather}} 
& 96 & \textbf{0.149} & \textbf{0.199} & 0.155 & 0.208 & 0.153 & 0.207 & 0.194 & 0.256 \\
& 192 & \textbf{0.193} & \textbf{0.241} & 0.194 & \textbf{0.242} & 0.194 & 0.244 & 0.230 & 0.283 \\
& 336 & \textbf{0.234} & \textbf{0.279} & 0.244 & 0.283 & 0.247 & 0.288 & 0.257 & 0.305 \\
& 720 & 0.308 & \textbf{0.329} & \textbf{0.307} & 0.330 & \textbf{0.307} & 0.330 & 0.326 & 0.350 \\
\bottomrule
\end{tabular}
}
\vspace{-0.1in}
\end{table*}

\subsection{Datasets}
We evaluate long-term forecasting performance of our proposed \MoU~on 7 commonly used datasets~\cite{wu2021autoformer}, including Weather, ILI and four ETT datasets (ETTh1, ETTh2, ETTm1 and ETTm2). Details are in Appendix A.
\subsection{Baselines and Setup}

To evaluate the overall performance of \MoU~in time series long-term forecasting task, we conducted a comparative analysis against four state-of-art models, each representative of a distinct architectural paradigm. 
The included baselines are as follows:
\begin{itemize}
    \item Mamba-based Models ({\bf S-Mamba}): S-Mamba model, known for reducing computational cost from quadratic to linear while maintaining competitive performance, has been effectively applied to time series forecasting, making it a relevant baseline for our study.
    \item Linear-based Models ({\bf D-Linear}): The D-Linear model, which challenges the dominance of Transformer-based models by achieving superior performance in time series forecasting, represents our chosen baseline for Linear-based models due to its simplicity and efficiency.
    \item Convolution-based Models ({\bf ModernTCN}): by expanding receptive field, demonstrates that convolutional models can achieve state-of-the-art performance in long-term time series forecasting, making it our selected baseline for Convolution-based models.
    \item Transformer-based Models ({\bf PatchTST}): PatchTST leverages the superiority of Transformer architecture meanwhile addressing its quadratic complexity by patching, making it our chosen representative for handling long-term dependencies in time series forecasting.
  
\end{itemize}

In our study, all models follow the same experimental setup with prediction length \( T \in \{24, 36, 48, 60\} \) for ILI dataset and \( T \in \{96, 192, 336, 720\} \) for other datasets as in PatchTST(~\cite{nie2022time}). 
We rerun the baseline models for four different look-back window \( L \in \{48, 60, 104, 144\} \) for ILI and \( L \in \{96, 192, 336, 720\} \) for others, and always choose the best results to create strong baselines. More details can be found in Appendix B.

\subsection{Main Results}

Multivariate long-term forecasting results are shown in Table \ref{t:main_result}. Overall, \MoU~consistently outperforms other baselines. Compared to ModernTCN, \MoU~achieves a 17.2\% reduction in MSE and a 9.5\% reduction in MAE. When compared to PatchTST, \MoU~yields a 22.6\% reduction in MSE and a 15.6\% reduction in MAE. Additionally, \MoU~demonstrates significant improvements across all datasets over the Linear-based DLinear and Mamba-based S-Mamba models.

\begin{table*}[t]
\centering
\caption{Ablation experiments for the design of long-term encoders with various architectures. It is noted that the order of layers significantly influence the performance. Best results are shown in \textbf{bold}.}
\label{t:ablation2}
\vspace{-0.1in}
\resizebox{0.8\textwidth}{!}{
\begin{tabular}{cc|cc|cc|cc|cc|cc|cc}
\toprule
\multicolumn{2}{c|}{Model}& \multicolumn{2}{c|}{MoA (in \MoU~)} & \multicolumn{2}{c|}{AA} & \multicolumn{2}{c|}{MM} & \multicolumn{2}{c|}{MFA} & \multicolumn{2}{c|}{AAA}& \multicolumn{2}{c}{MMA}\\ 
\hline
\multicolumn{2}{c|}{Metric} 
& \multicolumn{1}{c}{MSE} & \multicolumn{1}{c|}{MAE} 
& \multicolumn{1}{c}{MSE} & \multicolumn{1}{c|}{MAE}
& \multicolumn{1}{c}{MSE} & \multicolumn{1}{c|}{MAE}
& \multicolumn{1}{c}{MSE} & \multicolumn{1}{c|}{MAE}
& \multicolumn{1}{c}{MSE} & \multicolumn{1}{c|}{MAE}
& \multicolumn{1}{c}{MSE} & \multicolumn{1}{c}{MAE}\\
\hline
 \multirow{4}{*}{\rotatebox{90}{ETTh1}} 
& 96 & \textbf{0.358} & \textbf{0.393} & 0.370 & 0.396 & 0.372 & 0.394 & 0.367 & \textbf{0.393} & 0.376 & 0.401 & 0.369 & 0.395 \\
& 192 & \textbf{0.402} & 0.418 & 0.411 & 0.419 & 0.413 & 0.416 & 0.407 & \textbf{0.415} & 0.414 & 0.421 & 0.414 & 0.419 \\
& 336 & \textbf{0.389} & 0.418 & 0.390 & \textbf{0.412} & 0.413 & 0.430 & 0.404 & 0.420 & 0.416 & 0.432 & 0.422 & 0.435 \\
& 720 & \textbf{0.440} & 0.462 & 0.456 & 0.464 & 0.468 & 0.473 & 0.471 & 0.463 & 0.479 & 0.479 & 0.456 & \textbf{0.457} \\
 \hline
 \multirow{4}{*}{\rotatebox{90}{ETTm2}} 
& 96 & \textbf{0.166} & 0.257 & 0.168 & 0.257 & 0.169 & 0.259 & 0.169 & 0.258 & \textbf{0.166} & \textbf{0.256} & 0.167 & 0.257 \\
& 192 & \textbf{0.220} & \textbf{0.294} & 0.222 & \textbf{0.294} & 0.222 & 0.295 & 0.223 & 0.296 & 0.223 & 0.296 & 0.224 & 0.297 \\
& 336 & \textbf{0.270} & \textbf{0.329} & 0.281 & 0.333 & 0.281 & 0.335 & 0.281 & 0.334 & 0.275 & 0.330 & 0.275 & 0.331 \\
& 720 & \textbf{0.351} & \textbf{0.380} & 0.365 & 0.386 & 0.355 & 0.381 & 0.356 & 0.384 & 0.359 & 0.383 & 0.356 & 0.383 \\
 \hline
 \multirow{4}{*}{\rotatebox{90}{Weather}} 
& 96 & 0.149 & \textbf{0.199} & 0.152 & 0.203 & 0.156 & 0.206 & 0.150 & 0.201 & 0.151 & 0.202 & 0.152 & 0.202 \\
& 192 & 0.193 & \textbf{0.241} & 0.194 & 0.243 & 0.201 & 0.248 & 0.195 & 0.244 & 0.195 & 0.245 & 0.193 & 0.242 \\
& 336 & \textbf{0.234} & \textbf{0.279} & 0.242 & 0.283 & 0.243 & 0.286 & 0.245 & 0.284 & 0.244 & 0.284 & 0.244 & 0.283 \\
& 720 & 0.308 & \textbf{0.329} & \textbf{0.308} & 0.331 & 0.316 & 0.334 & 0.310 & 0.333 & 0.309 & 0.331 & 0.309 & 0.332 \\
\bottomrule
\end{tabular} 
}
\centering
\resizebox{0.8\textwidth}{!}{
\begin{tabular}{cc|cc|cc|cc|cc|cc|cc}
\toprule
\multicolumn{2}{c|}{Model}& \multicolumn{2}{c|}{MoA (in \MoU~)} & \multicolumn{2}{c|}{AMM} & \multicolumn{2}{c|}{MAM} & \multicolumn{2}{c|}{AMA} & \multicolumn{2}{c|}{AFM}& \multicolumn{2}{c}{AFCM}\\ 
\hline
\multicolumn{2}{c|}{Metric} 
& \multicolumn{1}{c}{MSE} & \multicolumn{1}{c|}{MAE} 
& \multicolumn{1}{c}{MSE} & \multicolumn{1}{c|}{MAE}
& \multicolumn{1}{c}{MSE} & \multicolumn{1}{c|}{MAE}
& \multicolumn{1}{c}{MSE} & \multicolumn{1}{c|}{MAE}
& \multicolumn{1}{c}{MSE} & \multicolumn{1}{c|}{MAE}
& \multicolumn{1}{c}{MSE} & \multicolumn{1}{c}{MAE}\\
\hline
 \multirow{4}{*}{\rotatebox{90}{ETTh1}} 
& 96 & \textbf{0.358} & \textbf{0.393} & 0.372 & 0.394 & 0.365 & 0.394 & 0.368 & 0.395 & 0.369 & 0.394 & 0.367 & \textbf{0.393} \\
& 192 & \textbf{0.402} & 0.418 & 0.420 & 0.422 & 0.411 & 0.418 & 0.413 & 0.420 & 0.410 & 0.417 & 0.410 & 0.417 \\
& 336 & \textbf{0.389} & 0.418 & 0.429 & 0.440 & 0.423 & 0.438 & 0.426 & 0.438 & 0.407 & 0.423 & 0.397 & 0.417 \\
& 720 & \textbf{0.440} & 0.462 & 0.474 & 0.475 & 0.470 & 0.464 & 0.587 & 0.515 & 0.465 & 0.467 & 0.465 & 0.470 \\
 \hline
 \multirow{4}{*}{\rotatebox{90}{ETTm2}}
& 96 & \textbf{0.166} & 0.257 & 0.168 & 0.257 & 0.167 & 0.257 & 0.169 & 0.257 & 0.168 & 0.257 & 0.168 & 0.257 \\
& 192 & \textbf{0.220} & \textbf{0.294} & 0.226 & 0.298 & 0.223 & 0.295 & \textbf{0.220} & \textbf{0.294} & 0.225 & 0.296 & 0.224 & 0.296 \\
& 336 & \textbf{0.270} & \textbf{0.329} & 0.278 & 0.331 & 0.277 & 0.331 & 0.276 & 0.330 & 0.281 & 0.333 & 0.277 & 0.330 \\
& 720 & \textbf{0.351} & \textbf{0.380} & 0.356 & 0.381 & 0.365 & 0.386 & 0.358 & 0.385 & 0.356 & \textbf{0.380} & 0.357 & \textbf{0.380} \\

 \hline
 \multirow{4}{*}{\rotatebox{90}{Weather}} 
& 96 & 0.149 & \textbf{0.199} & 0.157 & 0.208 & \textbf{0.148} & \textbf{0.199} & 0.152 & 0.203 & 0.151 & 0.202 & 0.150 & 0.200 \\
& 192 & 0.193 & \textbf{0.241} & 0.195 & 0.244 & \textbf{0.193} & 0.242 & \textbf{0.192} & 0.243 & 0.196 & 0.245 & 0.193 & 0.242 \\
& 336 & \textbf{0.234} & \textbf{0.279} & 0.251 & 0.287 & 0.247 & 0.284 & 0.242 & 0.280 & 0.236 & 0.280 & 0.236 & 0.282 \\
& 720 & 0.308 & \textbf{0.329} & 0.315 & 0.333 & 0.318 & 0.334 & \textbf{0.307} & 0.331 & 0.309 & 0.330 & 0.313 & 0.332 \\
\bottomrule
\end{tabular}
}
\end{table*}

\subsection{Ablation Study}
\label{sec:ablation}

\subsubsection{Ablation for feature extractor design.}
In this section, we investigate the efficacy of MoF in capturing short-term dependencies within patch tokens. We compare four types of encoders: one using a uniform transformation approach ({\bf Linear}) and three employing adaptive transformation methods, including our proposed {\bf MoF}, \textbf{SE-M}\footnote{SE-M is a modified method based on Squeeze-and-Excitation \cite{hu2019squeeze}, which applies an element-wise multiplication on the linearly transformed representation by a gating vector.}, and \textbf{Dyconv}~\cite{chen2020dynamic}. All methods take input $X_p\in\mathbb{R}^{N\times P}$ and produce output $X_\text{rep}\in\mathbb{R}^{N\times D}$. 

From Table \ref{t:abation1}, we make the following observations:

\begin{itemize}
    \item MoF outperforms the uniform transformation method (Linear), demonstrating that adaptive feature extraction leads to better representation of patch tokens.
    \item Dyconv, another adaptive method, fails to outperform Linear. We attribute Dyconv's poor performance to the huge increase in network parameters, making it unsuitable for small datasets like time series patches. This highlights our MoF's advantage in maintaining a small set of active parameters, achieving both strong performance and high efficiency.
    \item MoF also outperforms SE-M on most datasets. MoF’s diverse sub-extractors generate varied representations in different contexts, whereas SE-M's calibration strategy, which multiplies the original representation by a normalized gating vector, limits its ability of handling highly diverse contextual information.
    
\end{itemize}
In summary, MoF shows superior performance across all datasets, which demonstrates the effectiveness of our design in capturing short-term dependencies.

\subsubsection{Ablation for long-term encoders design.}
In this section, we focalize our ablation study on the MoA design. We design 10 baselines: AA, MM, MFA, AAA, MMA, AMM, MAM, AMA, AFM, and AFCM for comprehensive comparison. Wherein, A, M, F, and C represent Self-Attention, Mamba, FeedForward, and Convolution-layers, respectively, with the letter order indicating the sequence of layers. The experimental results are presented in Table \ref{t:ablation2}.

First, we examine the impact of the order between Mamba (M) and Self-Attention (A) layers on model performance. Models with the M-A order (MAM, AMA, MMA) consistently outperform those without it (AMM), suggesting that placing Mamba before Self-Attention is more effective for capturing long-term dependencies. Conversely, the model without the A-M order, MMA, still performs well compared to those with A-M (MAM, AMA, AMM), indicating that A-M is less effective than M-A for enhancing performance.

These findings highlight the significance of the Mamba and Self-Attention layer order, with the M-A order proving superior. We attribute this to the gradual expansion of long-term dependency awareness, from the partial view in the Mamba-layer to the global view in the Self-Attention layer, which enhances comprehensive long-term dependency learning. In Section \ref{sec:model_analysis}, we study deeper into what the Mamba and Self-Attention layers specifically capture.

Second, we explore the design of the transition layer between Mamba and Self-Attention layers. We compare two structures: a single FeedForward-layer (F) and a combination of FeedForward and Convolution-layers (F-C). Results show that the F-C transition (our MoA) significantly outperforms the F transition (MFA), demonstrating the effectiveness of F-C. The Convolution-layer further expands the receptive field of the Mamba-layer, providing an intermediate view that bridges the Mamba-layer’s partial view and the Self-Attention layer's global view.

\subsection{Model Analysis}
\label{sec:model_analysis}
\subsubsection{\textbf{Does MoF actually learn contexts within patches?}} 
\begin{figure}[t]
    \centering
    \includegraphics[width=0.99\linewidth]{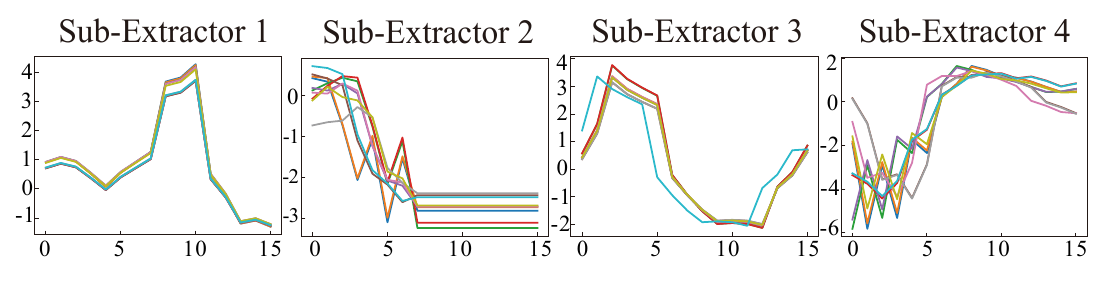}
    \vspace{-0.22in}
    \caption{The patches categorized by their activated sub-extractors automatically. }
    \label{fig:experts}
    \vspace{-0.15in}
\end{figure}

To assess whether MoF effectively learns distinct contexts within patches, we analyze the sub-extractor activations. After training MoF, we input a set of patches and record the activated sub-extractor for each. In cases where multiple sub-extractors are activated, we noted the one with the highest score, as detailed in Section \ref{sec:mof}. We then categorize the patches into $C$ classes based on their sub-extractor activations. As shown in Figure \ref{fig:experts}, patches associated with the same sub-extractor exhibit similar wave patterns, while those linked to different sub-extractors show divergent dynamics. This confirms that MoF effectively learns distinct contexts within patches, leading to more representative embeddings. More details and more visualizations can be found in Appendix E.

\begin{figure}[t]
    \centering
    \includegraphics[width=0.99\linewidth]{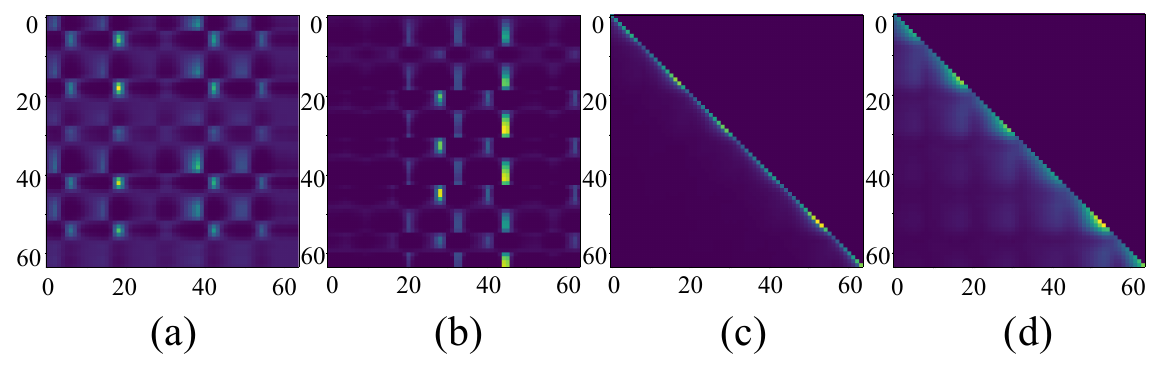}
    \vspace{-0.12in}
    \caption{Visualization of hidden attention maps of Self-Attention Layer (a-b) and Mamba-Layer(c-d) in MoA.}
    \label{fig:att_map}
    \vspace{-0.15in}
\end{figure}

\subsubsection{\textbf{What is learned by the layers of MoA?}}
As aforementioned, MoA is designed to capture long-term dependencies by initially focusing on partial dependencies in the Mamba-layer and gradually expanding to a global perspective in the Self-Attention layer. 
To validate this design, we analyze the attention maps of both the Mamba-layer and the Self-Attention layer. Since the Mamba-layer does not naturally generate attention maps, we use a method referenced in ~\cite{ali2024hidden} to create one. Figure \ref{fig:att_map} shows these attention maps, with panels (a) and (b) for Self-Attention layer and panels (c) and (d) to Mamba-layer.

In panels (c) and (d), the diagonal brightness gradient indicates that the Mamba-layer primarily captures recent long-term dependencies. The semi-bright grids in panel (d), starting from the diagonal and gradually fading, suggest that the Mamba-layer detects periodic long-term dependencies within a limited view. Conversely, panels (a) and (b) show scattered bright pixels, indicating that the Self-Attention layer captures long-term dependencies from a global perspective. 
These findings align with our initial design, confirming the rationale and effectiveness of our layer arrangement in MoA.

\section{Related Work}
Research into architectures for long-term time series forecasting recently has gained significant attention. For instance, ETSformer~\cite{woo2022etsformer} proposes time series Transformers by integrating the principle of exponential smoothing. PatchTST~\cite{nie2022time} introduces a Transformer-based model using patching and channel-independent structures for time series forecasting. ModernTCN~\cite{donghao2024moderntcn} optimizes the use of convolution in time series by proposing a DWConv and ConvFFN-based structure. TimeMachine~\cite{ahamed2024timemachine} leverages Mamba, a state-space model, to capture long-term dependencies in multivariate time series while maintaining linear scalability and low memory usage. Mambaformer~\cite{xu2024integrating} combines Transformer and Mamba architectures for enhanced time series forecasting. Additionally, Time-LLM~\cite{wang2023timemixer} repurposes large language models (LLMs) of LLaMA~\cite{touvron2023llama} for general time series forecasting by using LLMs as a reprogramming framework while keeping the backbone language models intact.

\section{Conclusion}

We have presented Mixture of Universals (\MoU), a pioneering and advanced model designed for efficient and effective time series forecasting. \MoU~consists of two key components: \MoFfull~(MoF), an adaptive method specifically designed to enhance time series patch representations for capturing short-term dependencies, and \MoAfull~(MoA), which hierarchically integrates multiple architectures in a structured sequence to model long-term dependencies from a hybrid perspective. The proposed approach achieves state-of-the-art performance across various real-world benchmarks while maintaining low computational costs. We hope that our work can bring new ideas and insights to the model structure design in the field of time series forecasting.

\bibliography{anonymous-submission-latex-2025}

\newpage
\clearpage

\section*{\LARGE{Appendix}}

\renewcommand{\thesection}{\Alph{section}}
\setcounter{section}{0}
\noindent In this Appendix, we provide details which are omitted in the main text. The outlines are as follows:
\begin{itemize}
    \item Section~\ref{Datasets Details}: An introduction of datasets and their training set split configurations (in ``Experiments" of main text).
    \item Section~\ref{Implementation Details}: A description of training configurations, including devices, optimization, training parameter settings and the visualization of predictions compared with baseline models (in ``Experiments" of main text).
    \item Section~\ref{Model Architecture and Parameter Details}: The details about model parameters configuration in our \MoU~(in ``Experiments" of main text).
    \item Section~\ref{Baseline Models Details}: An introduction to the baseline models and an elucidation of their respective calculation processes (in ``Ablation Study" of main text).
    \item Section~\ref{More Model Analysis}: More details for behaviors of MoF, and a discussion about differences between MoF and other similar existing works (in ``Model Analysis" of main text).
    \item Section~\ref{Metrics Illustration}: An illustration of metrics used to evaluate all models (in ``Experiments" of main text).
    \item Section~\ref{More Experiments}: The performance of \MoU~for univariate long-term forecasting, and analysis for impact of various look-back windows (in ``Main Results" of main text).   
    \item Our code as well as the running scripts are available at \color{blue}{\url{https://github.com/lunaaa95/mou/}}.
    
\end{itemize}

\section{Datasets Details}
\label{Datasets Details}
We evaluate the performance of \MoU~on seven widely used datasets~\cite{wu2021autoformer}, including Weather, ILI, Electricity and four ETT datasets (ETTh1, ETTh2, ETTm1 and ETTm2). All datasets are publicly available. 
\begin{itemize}
    \item \textbf{ETT datasets}\footnote{https://github.com/zhouhaoyi/ETDataset} contain two-year records of Electricity Transformer Temperature from two distinct regions in a Chinese province, labeled with suffixes 1 and 2. Each dataset includes seven variables with timestamps: HUFL, HULL, MUFL, LUFL, LULL, and OT. For univariate forecasting, OT (oil temperature) is the primary focus. The datasets are provided in two time resolutions: hourly (`h') and 15-minute intervals (`m').
    \item \textbf{Weather}\footnote{https://www.bgc-jena.mpg.de/wetter/} comprises 21 meteorological indicators from Germany for the year 2020, including variables like humidity, air temperature, recorded at 10-minute intervals.
    \item \textbf{ILI}\footnote{https://gis.cdc.gov/grasp/fluview/fluportaldashboard.html} is a national dataset tracking influenza-like illness, including patient counts and the illness ratio. It contains 7 variables, with data collected weekly.
    \item \textbf{Electricity}\footnote{https://archive.ics.uci.edu/ml/datasets/-\\ElectricityLoadDiagrams20112014} dataset records the hourly electricity consumption of 321 customers.
\end{itemize}

Dataset details are presented in Table~\ref{t:dataset}. Following the setup in \cite{nie2022time}, we split each dataset into training, validation, and test sets. The split ratio is 6:2:2 for ETT datasets and 7:1:2 for the others. The best parameters are selected based on the lowest validation loss and then applied to the test set for performance evaluation.

\begin{table*}[!t]
\centering
\caption{Description of seven commonly used datasets for time series forecasting.}
\label{t:dataset}
\begin{tabular}{c|ccccccc}
\toprule
Datasets & ETTh1 & ETTh2 & ETTm1 & ETTm2 & Weather & ILI & Electricity \\ \midrule
Number of Variables & 7 & 7 & 7 & 7 & 21 & 7 & 321 \\
Number of Timesteps & 17420 & 17420 & 69680 & 69680 & 52696 & 966 & 26304 \\
\bottomrule
\end{tabular}
\end{table*}

\section{Implementation Details}
\label{Implementation Details}

\subsubsection{Devices} All the deep learning networks are implemented in PyTorch and conducted on NVIDIA V100 32GB GPU. 
\subsubsection{Optimization} Our training target is minimizing $l_2$ loss, with optimizer of Adam by default. Then initial training rate is set to $2.5\times10^{-3}$ on ILI dataset for quick searching, $2\times10^{-4}$ for ETTm2 for better convergence, and $1\times10^{-3}$ for other datasets by default.
\subsubsection{Parameter Setting} Instead of using a fixed look-back window, we rerun PatchTST, ModernTCN, and S-Mamba with varying look-back windows: $L\in\{48, 60, 104, 144\}$ for the ILI dataset and $L\in\{192, 336, 512, 720\}$ for the other datasets. The best results are selected to generate strong baselines. For DLinear, we directly use the results from PatchTST~\cite{nie2022time}, where the best results with varying look-back windows are already selected.

\subsubsection{Visualization of prediction}
\begin{figure}[h]
    \centering
    \vspace{-0.1in}
    \includegraphics[width=0.99\linewidth]{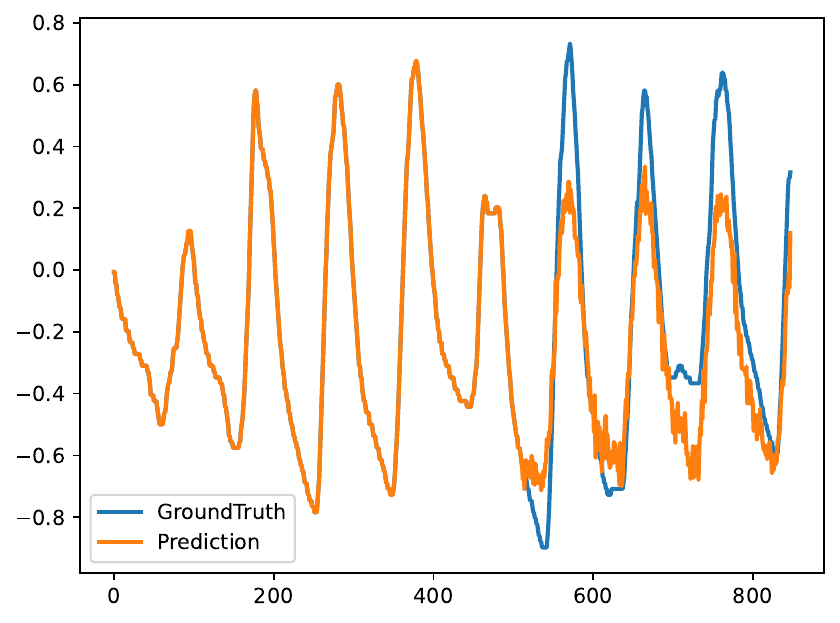}
    \vspace{-0.12in}
    \caption{Visualization of~\MoU's predicted values with ground truth.}
    \label{fig:ettm2_512_336}
    \vspace{-0.15in}
\end{figure}

We present a visualization of the future values predicted by \MoU~with the ground truth values for the ETTm2 dataset. The look-back window is set to 512 and predicted length is set to 336. 
The results are shown in Figure \ref{fig:ettm2_512_336}). 
We observe that the predicted values (orange line) are highly consistent with ground-truth (blue line), indicating our model is capable of making accurate forecasting. Besides, we also notice that the prediction repeats the periodic waves which have shown in historical series, indicating a long-term dynamics captured by model.

\begin{figure}[h]
    \centering
    \vspace{-0.1in}
    \includegraphics[width=0.99\linewidth]{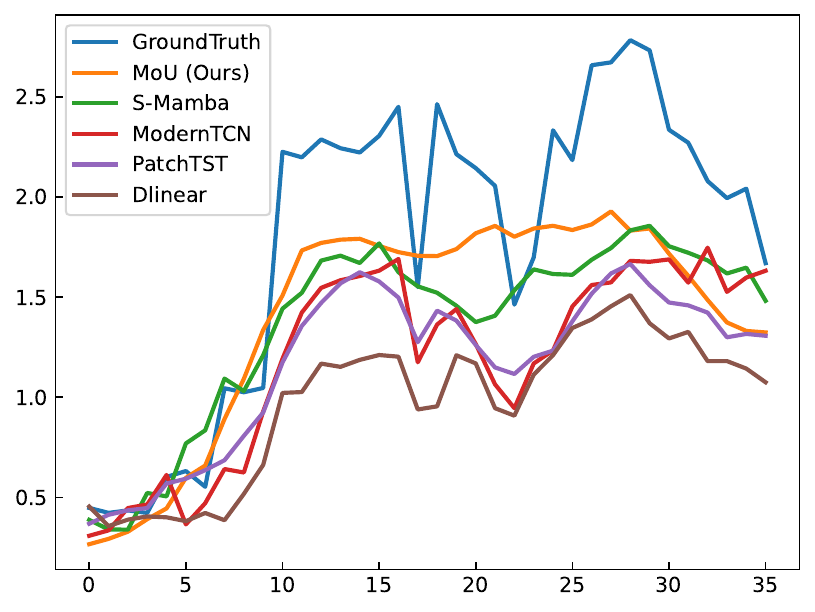}
    \vspace{-0.12in}
    \caption{Visualization of~\MoU's predicted values in comparison with other baseline models.}
    \label{fig:ILI-pd-gt}
    \vspace{-0.15in}
\end{figure}

We further collect the predictions of~\MoU, PatchTST~\cite{nie2022time}, Dlinear~\cite{zeng2023transformers}, ModernTCN~\cite{donghao2024moderntcn}, S-Mamba~\cite{wang2024mamba} on ILI dataset with a forecast horizon set to 36. The comparison results are displayed on \ref{fig:ILI-pd-gt}. The actual future values are denoted by the blue line, representing the ground truth.
Our proposed model, \MoU, is depicted by the orange line and is observed to most closely align with the ground truth, indicating its best accuracy in comparison with other baseline models.

\section{Model Architecture and Parameter Details}
\label{Model Architecture and Parameter Details}
By default, our \texttt{\textbf{MoU}} model includes a single MoA block. Our experimental setup uses a uniform patch length of 16 and a patch stride of 8 for generating patch tokens. The number of Sub-Extractors is set to four, with the top two selected in the MoF for generating patch representations. The Mamba layer employs an SSM state expansion factor of 21. In the FeedForward layer, the hidden size expansion rate is set to 2. The Convolution layer parameters are fixed with a kernel size of 3, a stride of 1, and padding of 1. In the Self-Attention layer, the number of heads is set to 4 for the ETTh1, ETTh2, and ILI datasets, and 16 for all other datasets.

The dimensionality of both short-term and long-term representations is set to 64 for the ETTh1 and ETTh2 datasets and increased to 128 for the remaining datasets for better performance. Within the MoA block, the input and output dimensions are kept consistent across layers, but can be adjusted by modifying the FeedForward layer's output dimension or the Convolution layer's kernel size and stride.

\begin{figure*}[t]
    \centering
    \includegraphics[width=0.95\linewidth]{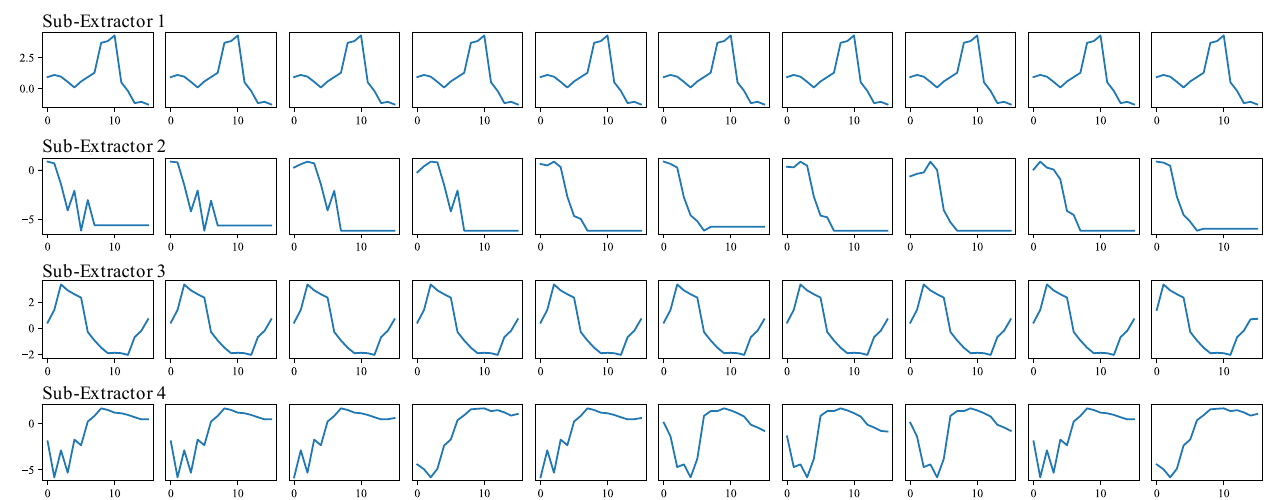}
    \vspace{-0.12in}
    \caption{The patches categorized by their activated Sub-Extractors in MoF. Separately displayed.}
    \label{fig:sub-extractors-sep}
    \vspace{-0.15in}
\end{figure*}

\begin{table*}[t]
\centering
\caption{Univariate long-term forecasting results of \MoU~on ETTh1, ETTh2 and ETTm1. Following setup of PatchTST~\cite{nie2022time}, look-back window is fixed to 336 with predicted lengths \( T \in \{96, 192, 336, 720\} \). Best results are shown in \textbf{bold}.}
\label{t:uni}
\begin{tabular}{cc|cc|cc|cc|cc}
\toprule
\multicolumn{2}{c|}{Model} & \multicolumn{2}{c|}{\texttt{\textbf{MoU}} (Ours)} & \multicolumn{2}{c|}{ModernTCN} & \multicolumn{2}{c|}{PatchTST} & \multicolumn{2}{c}{DLinear} \\ \hline
\multicolumn{2}{c|}{Metric} & \multicolumn{1}{c}{MSE} & \multicolumn{1}{c|}{MAE} & \multicolumn{1}{c}{MSE} & \multicolumn{1}{c|}{MAE} & MSE & MAE & MSE & MAE \\ 

\hline
\multirow{4}{*}{\rotatebox{90}{ETTh1}} 
& 96 & \textbf{0.052} & \textbf{0.176} & 0.055 & 0.179 & 0.055 & 0.179 & 0.056 & 0.180 \\
& 192 & \textbf{0.065} & \textbf{0.199} & 0.070 & 0.205 & 0.071 & 0.205 & 0.071 & 0.204 \\
& 336 & \textbf{0.073} & \textbf{0.214} & 0.074 & \textbf{0.214} & 0.081 & 0.225 & 0.098 & 0.244 \\
& 720 & \textbf{0.085} & \textbf{0.230} & 0.086 & 0.232 & 0.087 & 0.232 & 0.189 & 0.359 \\
\hline
 
\multirow{4}{*}{\rotatebox{90}{ETTh2}} 
& 96 & 0.126 & 0.278 & \textbf{0.124} & \textbf{0.274} & 0.129 & 0.282 & 0.131 & 0.279 \\
& 192 & \textbf{0.155} & \textbf{0.316} & 0.164 & 0.321 & 0.168 & 0.328 & 0.176 & 0.329 \\
& 336 & 0.173 & 0.341 & \textbf{0.171} & \textbf{0.336} & 0.185 & 0.351 & 0.209 & 0.367 \\
& 720 & \textbf{0.198} & \textbf{0.361} & 0.228 & 0.384 & 0.224 & 0.383 & 0.276 & 0.426 \\
\hline
 
\multirow{4}{*}{\rotatebox{90}{ETTm1}} 
& 96 & \textbf{0.026} & 0.123 & \textbf{0.026} & \textbf{0.121} & \textbf{0.026} & \textbf{0.121} & 0.028 & 0.123 \\
& 192 & \textbf{0.039} & \textbf{0.150} & 0.040 & 0.152 & \textbf{0.039} & \textbf{0.150} & 0.045 & 0.156 \\
& 336 & \textbf{0.052} & \textbf{0.173} & 0.053 & \textbf{0.173} & 0.053 & \textbf{0.173} & 0.061 & 0.182 \\
& 720 & \textbf{0.073} & 0.209 & \textbf{0.073} & \textbf{0.206} & 0.074 & 0.207 & 0.080 & 0.210 \\

\midrule
\multicolumn{2}{c|}{Win count} & \multicolumn{2}{c|}{\color{red}{\textbf {18}}} & \multicolumn{2}{c|}{\underline{10}} & \multicolumn{2}{c|}{5} & \multicolumn{2}{c}{0}\\ 
\bottomrule
\end{tabular}
\end{table*}

\section{Baseline Models Details}
\label{Baseline Models Details}
We present the detailed calculations of SE-M, W, and Dyconv, which are baseline models in Section 3.4 (main text). 

\noindent\textbf{SE-M} is a modified Squeeze-and-Excitation~\cite{hu2019squeeze} method. The process of SE-M can be described as:
\begin{equation}
        {\bf X}_{rep}=\operatorname{SE-M}({\bf X}_p)= (W_{se} {\bf X}_p)\otimes \bf Y \\
\end{equation}
where parameter $W_{se}\in\mathbb{R}^{P\times D}$, and {\bf Y} denotes the gating vector, which can be describes as:
\begin{equation}
        {\bf Y}=\operatorname{Expand}(\sigma_2\left(W_2 \sigma_1\left(W_1 {\bf Z}\right)\right)) \\
\end{equation}
where $\sigma_1$ and $\sigma_2$ are respectively $\operatorname{ReLU}$ function and $\operatorname{Sigmoid}$ function. $\bf Z$ is a average pooled vector which squeezes the information:
\begin{equation}
        {\bf Z}=\operatorname{AvgPool}(W_{se}{\bf X}_p) \\
\end{equation}
To avoid underestimating the ability of original Squeeze-and-Excitation method, we apply Squeeze-and-Excitation on $W_{se}{\bf X}_p$ instead of vanilla ${\bf X}_p$ to generate enriched representation. This modification is the difference between original Squeeze-and-Excitation and our baseline method SE-M.
Overall, the parameters participated in once calculation are $W_{se}\in\mathbb{R}^{P\times D}$, $W_{1}\in\mathbb{R}^{D\times (D/r)}$ and $W_{2}\in\mathbb{R}^{(D/r)\times D}$, where $r$ is the reduction rate.

\textbf{W} is a linear transformation. The process can be summarized as:
\begin{equation}
    {\bf X}_{rep}=\operatorname{Linear}({\bf X}_p)=W_l {\bf X}_p
\end{equation}

\noindent\textbf{Dyconv} is the method of Dynamic Convolution~\cite{chen2020dynamic}. The process can be described as:
\begin{equation}
    {\bf X}_{rep}=\operatorname{Dyconv}({\bf X}_p)=\operatorname{Conv}({\bf X}_p;\mathcal{K})
\end{equation}
where $\mathcal{K}$ denotes the parameters of adaptive kernel, which is aggregated by a set of kernels $\{\mathcal{K}_i\}$ with attention weight of $\pi_i(X_p)$, this process can be described as:
\begin{equation}
    \mathcal{K}=\sum_{i=1}^{N} \pi_i({\bf X}_p) \mathcal{K}_i
\end{equation}
We calculate the weight of each kernel ${\pi_i({\bf X}_p)}$ by:
\begin{equation}
    {\pi_i({\bf X}_p)}=\operatorname{Softmax}(W_2(\operatorname{\sigma}(W_1(\operatorname{AvgPool}({\bf X}_p))))
\end{equation}
where $\sigma$ is activation function of $\operatorname{ReLU}$. The size of parameters in Dyconv is biggest among all listed methods.  
Overall, the parameters participated in once calculation are $W_{1}\in\mathbb{R}^{P\times D}$, $W_{2}\in\mathbb{R}^{D\times (D/r)}$ and $\{\mathcal{K}_i\}$.

\begin{table*}[t]
\centering
\caption{Impact of look-back windows to performance of \MoU~ for multivariate long-term forecasting. We test four look-back windows $L\in\{192, 336, 512, 720\}$. Best results are shown in \textbf{bold}.}
\label{t:look-back window}
\begin{tabular}{cc|cc|cc|cc|cc}
\toprule
\multicolumn{2}{c|}{$L$} & \multicolumn{2}{c|}{192} & \multicolumn{2}{c|}{336} & \multicolumn{2}{c|}{512} & \multicolumn{2}{c}{720} \\ \hline
\multicolumn{2}{c|}{Metric} & \multicolumn{1}{c}{MSE} & \multicolumn{1}{c|}{MAE} & \multicolumn{1}{c}{MSE} & \multicolumn{1}{c|}{MAE} & MSE & MAE & MSE & MAE \\ 

\hline
\multirow{4}{*}{\rotatebox{90}{ETTh1}} 
& 96 & 0.375 & 0.400 & \textbf{0.358} & \textbf{0.393} & 0.366 & 0.402 & 0.381 & 0.416 \\
& 192 & 0.418 & 0.423 & \textbf{0.402} & \textbf{0.418} & 0.435 & 0.446 & 0.429 & 0.445 \\
& 336 & 0.410 & 0.428 & \textbf{0.389} & \textbf{0.418} & 0.422 & 0.447 & 0.420 & 0.449 \\
& 720 & 0.488 & 0.488 & \textbf{0.440} & \textbf{0.462} & 0.461 & 0.479 & 0.477 & 0.494 \\
\hline
 
\multirow{4}{*}{\rotatebox{90}{ETTh2}} 
& 96 & 0.275 & 0.337 & 0.266 & 0.332 & \textbf{0.257} & \textbf{0.329} & 0.264 & 0.334 \\
& 192 & 0.338 & 0.378 & 0.319 & 0.369 & \textbf{0.312} & \textbf{0.367} & 0.323 & 0.375 \\
& 336 & 0.322 & 0.376 & 0.305 & 0.369 & \textbf{0.303} & \textbf{0.367} & 0.310 & 0.375 \\
& 720 & 0.395 & 0.428 & \textbf{0.379} & \textbf{0.422} & 0.382 & 0.427 & 0.389 & 0.433 \\
\hline
 
\multirow{4}{*}{\rotatebox{90}{ETTm2}} 
& 96 & 0.170 & \textbf{0.255} & 0.167 & 0.256 & 0.166 & 0.257 & \textbf{0.164} & 0.256 \\
& 192 & 0.228 & \textbf{0.294} & 0.220 & \textbf{0.294} & 0.220 & \textbf{0.294} & \textbf{0.218} & \textbf{0.294} \\
& 336 & 0.286 & 0.334 & 0.278 & 0.331 & 0.270 & 0.329 & \textbf{0.268} & \textbf{0.327} \\
& 720 & 0.368 & 0.383 & 0.359 & 0.384 & 0.351 & 0.380 & \textbf{0.346} & \textbf{0.377} \\
\bottomrule
 
\end{tabular}
\label{table5}
\end{table*}

\section{More Model Analysis}
\label{More Model Analysis}

\subsubsection{Behaviors of Sub-Extractors in MoF} We conducted an experiment to extract the contextual information learned by the Sub-Extractors in MoF. The method is descried in Section 3.5 (main text), with the experimental outcomes presented in an integrated format in Figure 5 (main text). Specifically, we selected the top 10 patches with the highest scores to serve as representative examples for their respective Sub-Extractors.

To enhance clarity, each patch is individually displayed in Figure \ref{fig:sub-extractors-sep}. Within each row of the figure, the patches are processed by the same Sub-Extractor. It is observable that the patches exhibit a consistent shape within rows, while there is a significant divergence in shapes across rows. This observation demonstrates the proficiency of the MoF in capturing contextual information.

\subsubsection{MoF vs. MoE} MoE has various applications. Originally, MoE is used to process data which naturally can be partitioned into multiple subsets by topic or domains~\cite{jacobs1991adaptive}. Then,~\cite{shazeer2017outrageously} reports that the application of MoE in Recurrent Neural Networks (RNN) to enlarge model capacity for very large datasets while maintaining efficiency. Further,~\cite{lepikhin2020gshard} extend the study of MoE to Transformers. Our MoF is based on MoE, and has similar structure with ~\cite{shazeer2017outrageously}. But there are still some differences. First, MoF is not designed for enlarging capacity for large datasets, but focus on small datasets with very divergent contexts (such as time series patches). Second, MoF is aimed to consider various semantic context hidden within time series patches, which have implicit meanings, while other applications of MoE are mainly designed for language modeling, which have explicit different meanings. Therefore, despite similar structures, MoF works for different purpose and processes different data type compared to existing MoE models.

\begin{figure}[h]
    \centering
    \vspace{-0.1in}
    \includegraphics[width=0.99\linewidth]{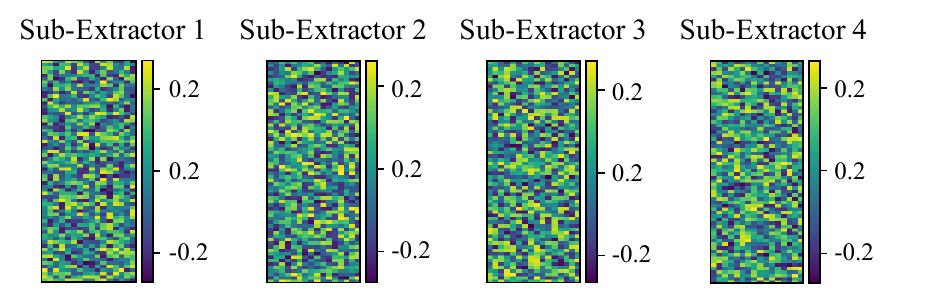}
    \vspace{-0.12in}
    \caption{Visualization of sub-extractor parameters.}
    \label{fig:expert_paras}
    \vspace{-0.15in}
\end{figure}

We also extract the parameters in Sub-Extractors on dataset ETTh1. The visualization of their linear matrices are displayed in Figure~\ref{fig:expert_paras}. Axis-0 denotes input dimension of patch, which is the same value with patch size, while axis-1 denotes output dimension set to 64 on ETTh1.

\section{Metrics Illustration}
\label{Metrics Illustration}
We use mean square error (MSE) and mean absolute error (MAE) as our metrics for evaluation of all forecasting models. Then calculation of MSE and MAE can be described as:
\begin{equation}
\hspace{0.3cm} \mathrm{MSE}=\frac{1}{T} \sum_{i=L+1}^{L+T}\left(\hat{\mathbf{X}}_i-\mathbf{X}_i\right)^2
\end{equation}

\begin{equation}
\mathrm{MAE}=\frac{1}{T} \sum_{i=L+1}^{L+T}\left|\hat{\mathbf{X}}_i-\mathbf{X}_i\right|
\end{equation}
where $\hat{\mathbf{X}}$ is predicted vector with $T$ future values, while $\mathbf{X}$ is the ground truth.

\section{More Experiments}
\label{More Experiments}

\subsubsection{Univariate long-term forecasting.} We also conduct univariate long-term forecasting experiments on four ETT datasets, focusing on oil temperature as the primary variable of interest. The results, presented in Table~\ref{t:uni}, demonstrate that our proposed \texttt{\textbf{MoU}} outperforms baseline models, including ModernTCN, PatchTST, and DLinear.

\subsubsection{Impact of input length}
The length of look-back window has significant impact on the performance of model, as longer look-back window indicates more historical information. However, information from too distant past may harm the accuracy of current predictions. Thus, we conduct experiments to evaluate~\MoU's performance with four look-back windows $L\in\{192, 336, 512, 720\}$, as shown in Table~\ref{t:look-back window}.

\end{document}